\documentclass[10pt, conference]{IEEEtran}

\ifCLASSINFOpdf
\else
\fi
\usepackage{amsmath}

\usepackage{svg}

\usepackage{url}

\usepackage{booktabs}

\begin{document}
\title{Comparison of machine learning models applied on anonymized data with different techniques}

\author{\IEEEauthorblockN{Judith Sáinz-Pardo Díaz}
\IEEEauthorblockA{Instituto de Física de Cantabria (IFCA), CSIC-UC\\
Avda. los Castros s/n. 39005 - Santander (Spain)\\
Email: sainzpardo@ifca.unican.es}
\and
\IEEEauthorblockN{Álvaro López García}
\IEEEauthorblockA{Instituto de Física de Cantabria (IFCA), CSIC-UC\\
Avda. los Castros s/n. 39005 - Santander (Spain)\\
Email: aloga@ifca.unican.es}
\and}

\maketitle

\begin{abstract}
Anonymization techniques based on obfuscating the quasi-identifiers by means of value generalization hierarchies are widely used to achieve preset levels of privacy. To prevent different types of attacks against database privacy it is necessary to apply several anonymization techniques beyond the classical \textit{k-anonymity} or \textit{$\ell$-diversity}. However, the application of these methods is directly connected to a reduction of their utility in prediction and decision making tasks. In this work we study four classical machine learning methods currently used for classification purposes in order to analyze the results as a function of the anonymization techniques applied and the parameters selected for each of them. The performance of these models is studied when varying the value of $k$ for \textit{k-anonymity} and additional tools such as \textit{$\ell$-diversity}, \textit{t-closeness} and \textit{$\delta$-disclosure privacy} are also deployed on the well-known \textit{adult dataset}.
\end{abstract}

\IEEEpeerreviewmaketitle

\section{Introduction} 
Digitization makes the amount of data being generated on a daily basis increasingly larger and more valuable for use and processing. The easier sharing of information with anyone else anywhere in the world has meant that the research and development of secure protocols for such releasing is proceeding apace. The continuous generation of individuals' personal data makes it essential to develop privacy preserving techniques and to include them in data science pipelines. 

The use of anonymization techniques is key in the publication, processing and analysis of sensitive data. This is a particularly critical issue in a data science process where we are dealing with data containing information about individuals that allow to identify them. 

To this aim, different privacy models have been proposed with the overall idea of providing data privacy and anonymity guarantees. When handling a database that is to be anonymized, one must clearly identify certain types of attributes or information in it according to their nature: \textit{identifiers} (variables which uniquely characterize an individual, e.g. ID number or name and surname), \textit{quasi-identifiers} (variables accessible to the attacker that, through their combination, make it possible to identify an individual, e.g. gender, age or address) and \textit{sensitive attributes} (confidential information about an individual that should not be associated with him/her, e.g. salary, clinical history, etc) \cite{sainz2022python}. Therefore, during an anonymization process, identifiers must be eliminated, and a taxonomy tree or scheme of hierarchies can be established for each quasi-identifier, allowing them to be generalized or even deleted \cite{domingo2016database}. All database records that are identical with respect to the quasi-identifiers form an \textit{equivalence class}. In this context, some classical anonymization techniques that focus both on acting on the quasi-identifiers (such us \textit{k-anonymity}) but also on the sensitive attributes and their distribution in the different equivalence classes of the database (such us \textit{$\ell$-diversity}, \textit{t-closeness} and \textit{$\delta$-disclosure privacy}) can be applied. 

The relevance of anonymization is self-evident in many areas where sensitive data are available and when machine learning (ML) models are potential tools to be applied in order to carry out an inference process with them. However, it should be noted that a too strict level of anonymity may compromise their usefulness for processing and inference. This can occur either because a large number of records had to be eliminated to reach the desired level of anonymization, or because the hierarchies applied on the quasi-identifiers dilute their initial information. It is necessary to achieve a balance between the level of data anonymization and the utility of the data for analysis. In this work we will use a classical dataset in the field of anonymization, together with some pre-established hierarchies on its quasi-identifiers, and we will anonymize it using four different methods. On these data, anonymized with different levels and different techniques, an inference process will be carried out using four classical machine learning models, from the classical \textit{k-Nearest Neighbors (kNN)} to three ensemble methods based on the use of decision trees, \textit{Random Forest (RF)}, \textit{Adaptive Boosting (AB)} and \textit{Gradient Tree Boosting (GB)}. 

The remainder of this paper is structured as follows: in Section~\ref{sec:related} the related work in the area is presented. In Section~\ref{sec:anonymity} the four anonymization techniques under study are exposed. Section~\ref{sec:results} presents the data used, the machine learning models employed, and the results obtained in each case according to different metrics. Finally, Section~\ref{sec:conclusions} draws the conclusions obtained from the study performed and Section~\ref{sec:code} presents the code availability. 

\section{Related work}\label{sec:related}

\subsection{Privacy preservation}\label{subsec:privacy}
\begin{figure}[ht]
    \centering
    \includegraphics[width = \linewidth]{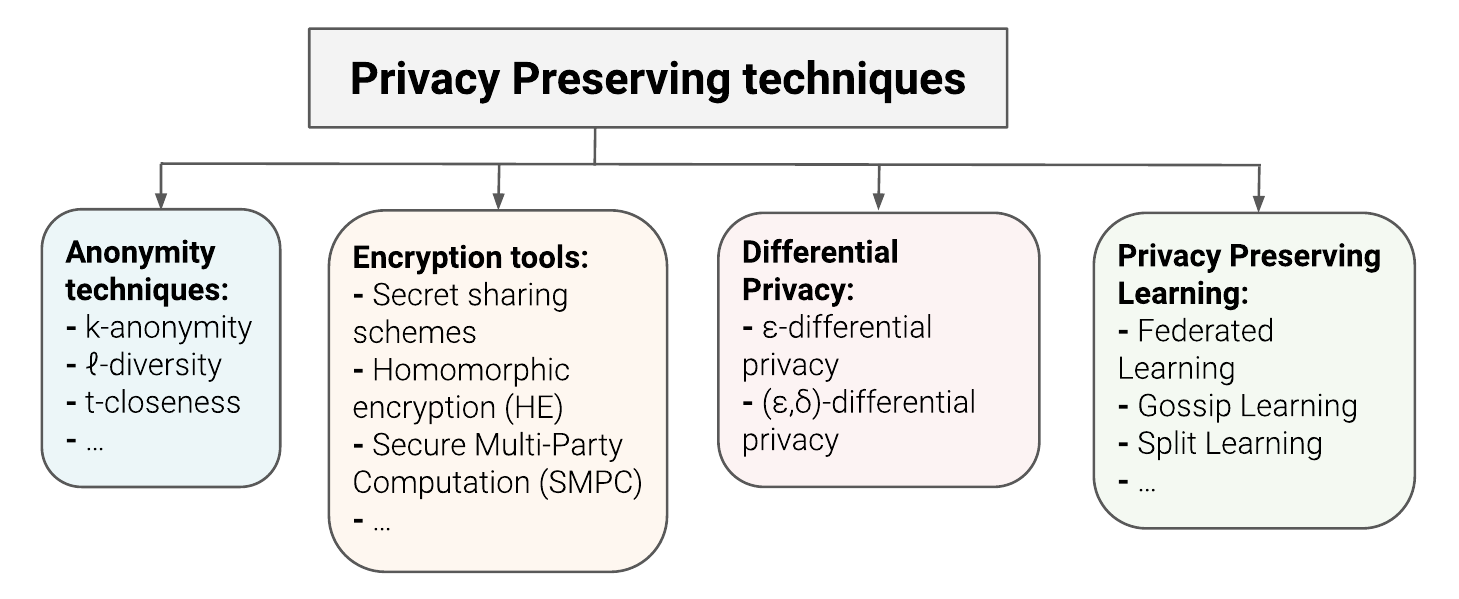}
    \caption{Privacy Preserving techniques. Inspired in Figure 2.1 of \cite{kim2021privacy}.}
    \label{fig:pptechniques}
\end{figure}

When managing data including sensitive attributes it is important to focus on data privacy in order to avoid possible security breaches, especially when these may involve a threat to an individual's privacy. To this end, privacy preserving technologies are in the foreground \cite{kim2021privacy}. Specifically, privacy-preserving technologies encompass a wide range of techniques that can be grouped into different subsets (see Figure~\ref{fig:pptechniques}). On the one hand, those that focus on cryptographic methods, such as Homomorphic Encryption (HE) or Secure Multi-Party Computation (SMPC), which make it possible to process and analyze the information without the need to decrypt it \cite{kim2021privacy}, thus preserving its integrity and privacy. On the other hand, techniques such as differential privacy are also emerging. The implementation of differential privacy is proposed in \cite{TCS-042}, and the main idea is to achieve that the absence of a single record does not impact the overall dataset characteristics. Several methods can be used in order to add such an statistical noise, e.g. Laplace and Exponential mechanisms among others \cite{zhu2017differential}. 
 
Other key privacy preserving technologies are in the focus of this work, namely anonymity techniques. Specifically, \textit{k-anonymity} was introduced in 1998 \cite{samarati1998protecting}, and seeks to make an individual in a dataset indistinguishable from at least $k-1$ other individuals \cite{kim2021privacy}. This approach is very simple but it is also widely used even nowadays (\cite{arava2020adaptive, li2013applications, 5365038}). However, it is susceptible to numerous attacks \cite{machanavajjhala2007diversity}, so other techniques must also be taken into account, such as \textit{$\ell$-diversity} or \textit{t-closeness} among many others, focusing on the sensitive attributes and their distribution in the database \cite{machanavajjhala2007diversity, li2006t}. Details about these techniques and the attacks they aim to prevent will be detailed in Section~\ref{sec:anonymity}. 

Finally, another group which could be included is \emph{privacy preserving machine/deep learning}. If we are focusing on data analysis by means of ML or DL models, data decentralization techniques, such as federated learning, split learning or gossip learning allow collaboration between different clients to apply this kind of models to their data without the need for them to share it with each other or with a central server \cite{SAINZPARDODIAZ2023142}.

\subsection{Data utility}\label{subsec:utility}
Here, the largest challenge that we find when applying anonymization techniques to preserve the privacy of a database lies in achieving a balance between the privacy of the data, and its usefulness for further analysis. There is no point in obtaining data with a level of anonymity that makes them insusceptible to any attack (which is an optimal scenario from a privacy point of view), if they are of no practical use at all. It is therefore essential to analyze the usefulness of the data after anonymization in order to select the level that best achieves this trade-off. To this end, in \cite{SLIJEPCEVIC2021102488} different \textit{k-anonymization} algorithms (Mondrian, OLA, TGD, CB) are studied together with the performance of four machine learning models by varying the value of $k$, concluding that with an increasingly strong \textit{k-anonymity} constraint, the classification performance generally degrades. Also, in this work the authors note that ``for very large $k$ of up to 100 the performance losses remain within acceptable limits''. On the other hand, and following this same line, in \cite{wimmer2014comparison} different artificial intelligence models (i.e. neural networks, logistic regression, decision trees, Bayesian classifier) are considered together with three datasets, in order to analyze the performance of the different models before and after applying \textit{k-anonymity} to the data with a value of $k=2$. The authors highlight that, according to their results, certain machine learning algorithms are more suited to be used with privacy-preserving data mining techniques than others. Finally, in \cite{goldsteen2022anonymizing} several examples applying both \textit{k-anonymity} and \textit{differential privacy} are presented in order to mitigate inference attacks.

Among our contributions we highlight the study of additional anonymization techniques to \textit{k-anonymity}, which focus on sensitive attributes. On the one hand, we first study the influence of different values of $k$ on the four analyzed machine learning models using two metrics: accuracy and AUC. On the other hand, we consider three other anonymity techniques and analyze their effect on the perfomance of the different models under study. For this purpose, we use a classical dataset and a set of predefined hierarchies to carry out the anonymization using the \textit{ARX Software} \cite{prasser2015putting}.

\section{Anonymization techniques deployed}\label{sec:anonymity}

In this study four classical anonymity techniques that focus on obfuscating quasi-identifiers by applying value generalization hierarchies (VGH) to them will be applied. Specifically, generalization consists of replacing the original values by other consistent but less specific ones, in this case by means of taxonomy tree or hierarchies. Maximal generalization or value suppression consists of replacing it with an special character such as \textit{`*'}.

Thus, we provide the dataset with a hierarchy scheme through which we can generalize the various quasi-identifiers to a more complete domain, as will be explained in the next section. Different examples of hierarchy trees can be found in Figure 1 of \cite{SLIJEPCEVIC2021102488} and in Figure 2 of \cite{ayala2014systematic} among others.

Specifically, we will start with the application of one of the most classic and widely used techniques, due to its easy interpretability and implementation, that is \textit{k-anonymtity}. A database is said to be \textit{k-anonymous} if and only if there are at least $k$ records in each equivalence class of the database. If this condition is verified the probability of that record corresponding to an individual is $1/k$. In the same way, another of the techniques applied, \textit{$\ell$-diversity}, is also widely used for preserving privacy of sensitive attributes of a database (especially to prevent homogeneity attacks), due again to its easy interpretability. We say that \textit{$\ell$-diversity} is verified if there are at least $\ell$ different values for the sensitive attribute in each equivalence class of the database \cite{machanavajjhala2007diversity}. Finally, two other techniques to be employed that focus on the distribution of sensitive attributes in the different equivalence classes of the database are \textit{t-closeness} and \textit{$\delta$-disclosure privacy}. A brief description of this last two methods is given below (see \cite{sainz2022python}).

\begin{itemize}
    \item \textit{t-closeness:} is verified if the distribution of the values of the sensitive attribute in each equivalence class are no more than a distance $t$ apart from the distribution of the sensitive attribute in the whole database. In order to measure the distance between the distributions, for numerical attributes the Earth Mover’s distance (EMD) with the ordered distance is applied, while for categorical attributes the equal distance is used \cite{li2006t}.
    \item \textit{$\delta$-disclosure privacy:} is satisfied if $\left|\log\left(\frac{p(EC,s)}{p(DB,s)}\right)\right|< \delta$ is fulfilled for every equivalence class (EC) and every value $s$ of the sensitive attribute, being $p(EC,s)$ its distribution in an equivalence class and $p(DB,s)$ in the whole database \cite{10.1145/1401890.1401904}. 
\end{itemize}

It is essential to study several anonymity techniques together, since each one of them can prevent a different type of attack on the database. Specifically, the above four techniques can be used respectively to prevent different kind of attacks, as summarized in Table~\ref{tab:anonymity_attacks} for the methods presented previously.

\begin{table}[ht]
    \caption{Main attacks that prevent each anonymity technique under study. Extracted from Table 2 of \cite{sainz2022python}.}
    \label{tab:anonymity_attacks}
    \centering
    \begin{tabular}{cl}
    \toprule
         \textit{\textbf{Technique}} & \textit{\textbf{Main attacks prevented}}\\
         \midrule
         \textit{k-anonymity} & Linkage and re-identification attacks \\
         \textit{$\ell$-diversity} & Homogeneity and background attacks\\
         \textit{t-closeness} & Similarity and skewness attacks\\
         \textit{$\delta$-disclosure privacy} & Skewness and inference attacks\\
    \bottomrule
    \end{tabular}
\end{table}

\section{Experimental results and analysis}\label{sec:results}
The data used in this analysis are presented below along with the hierarchies established for each quasi-identifier. In addition, a brief description of the machine learning models analyzed through this study is presented. The performance of these models will be analyzed in two cases: varying the value of $k$ fixed for \textit{k-anonymity}, and once anonymized with $k=5$, applying the other three techniques focused on the distribution of quasi-identifiers: \textit{$\ell$-diversity}, \textit{t-closeness} and \textit{$\delta$-disclosure privacy}.

\subsection{Dataset and hierarchies}\label{sec:data}

In order to carry out this comparative study, a classic dataset regarding privacy and anonymity techniques will be used: the \emph{adult dataset} (available in \cite{Dua2019}), which is an extraction of information from the 1994 U.S. Census database. A sub-sample containing 32561 records will be used, and the objective is to predict whether or not the income of an individual exceeds 50000$\$$ by year, based on census data.

Specifically in this study six quasi-identifiers (which would act as the features when training a machine learning model) will be considered, namely: \textit{age}, \textit{sex}, \textit{education}, \textit{relationship}, \textit{occupation} and \textit{native-country}, together with \textit{salary class} (which can take as values $>50K$ or $\leq50K$) acting as the sensitive attribute and the label of the data. 
The following hierarchies have been applied to each quasi-identifier (extracted from those presented in the \textit{ARX Software} GitHub repository \url{https://github.com/arx-deidentifier/arx}):
\begin{itemize}
    \item \textit{age:} five levels of hierarchies are applied, starting with 5-year intervals: $[15, 20)$, $[20, 25)$,$\hdots$, $[75, 80)$, and finally the case where the age is greater than 80. Further levels are defined with 10, 20, 40 and 80 years intervals. 
    \item \textit{sex:} no hierarchies are applied.
    \item \textit{education:} two levels of hierarchies are applied for each of the 16 possible values. In the first level there are five values: \textit{Primary School}, \textit{High School}, \textit{Undergraduate}, \textit{Professional Education} and \textit{Graduate}. The second level only takes three possible alternatives: \textit{Primary education}, \textit{Secondary education} and \textit{Higher education}.
    \item \textit{relationship:} no hierarchy is applied.
    \item \textit{occupation:} a hierarchy level is included that encompasses in three categories the different options, namely: \textit{technical}, \textit{non-technical} and \textit{other}.
    \item \textit{native country:} a hierarchy level is introduced that generalizes the country by continent. Thus, the possible options according to the countries present in the original database are: \textit{Africa}, \textit{Asia}, \textit{Europe}, \textit{North America}, \textit{South America} and \textit{unknown} (if the value of the native country field was ?). 
\end{itemize}
As already mentioned, the anonymization process will be carried out using the \textit{ARX Software} with the hierarchies described above for the quasi-identifiers and with \textit{salary class} as the sensitive attribute.

\subsection{Machine learning models and evaluation metrics}

Through this study we present a classical machine learning classification problem. That is, given a set of inputs, $\mathbf{X}\in \mathrm{D}^{n x m}$ with $n$ the number of records and $m$ the number of features, and the corresponding labels (outputs) $\mathbf{y}$, with $y_{i} \in \{1, \hdots, n_{C}\}$ $\forall i \in \{1,\hdots, n\}$, being $n_{C}$ the number of different classes, our objective is to estimate a function $\hat{f}$ which approximates $\mathbf{y}=\hat{f}(\mathbf{x})$ \cite{murphy2012machine}. Specifically, we are going to apply the following four supervised machine learning models, with the quasi-identifiers as features and the sensitive attribute as label:

\begin{itemize}
    \item \textit{\textbf{k-Nearest Neighbors (kNN):}} This non-parametric method find the $k$ nearest point in the training split to the test input. Thus, it computes the posterior probability that an element belongs to certain class. Its main drawback is that its performance is poor in cases of high dimensional inputs \cite{murphy2012machine}. While large values of $k$ reduce the effect of noise in classification, a value $k=1$ actually induces a Voronoi tessellation of the point $\mathbf{X}_{i}, i\in\{1,\hdots,n\}$.
    \item \textit{\textbf{Random Forest (RF):}} The main motivation for introducing this method comes from trying to reduce the variance of an estimator by aggregating several of them. In particular, $N$ decision trees can be trained and their ensemble be computed. In order to avoid obtaining highly correlated predictors, the Random Forest method learns trees based on randomly chosen subsets of the input data and of the features \cite{murphy2012machine}. Specifically, a new training subset is constructed using bootstrapping, then decision trees are trained on it using a subset of the predictor variables. This process is repeated to finally obtain a unified prediction.
    \item \textit{\textbf{Adaptive Boosting (AB):}} Boosting is a ML approach based on the idea of creating a highly accurate prediction rule by combining many relatively weak rules \cite{Schapire2013}. Thus, with adaptive boosting we first train a classifier (by giving equal weight to all data), calculate the error associated with it, and compute a new distribution to weight the data based on whether or not it was correctly classified. This process is repeated $M$ times and finally a weighted average of the classifiers is performed. The details of this method can be found in pseudocode form in Algorithm 16.2 of~\cite{murphy2012machine}. 
    \item \textit{\textbf{Gradient Tree Boosting (GB):}} This ensemble classifier trains several individual predictors sequentially. As in the case of Adaptive Boosting, again the fundamental idea lies in combining weak predictors (decision trees) to create a robust one. Typically, with this method the first predictor learns to predict the data mean, then the second one explains the errors of the first one, the third one explains the errors of the second one, and so on. A detailed formulation can be found in~\cite{friedman2002stochastic}.
\end{itemize}

The Python library \textit{scikit-learn} (version 1.2.0) has been used to train and test the different models. In addition, the \textit{GridSeachCV} function is used to select the optimal parameters for each of them in order to perform a 5-fold cross-validated grid-search. Then, the model is retrained with the optimum parameters calculated with the cross-validation. Specifically, in each model the hyper-parameter grid selected is presented bellow together with some other fixed parameters: 

\begin{itemize}
    \item \textbf{\textit{kNN.}} Number of neighbors: [3, 4, $\hdots$ , 50]. Metric to calculate the distance: \textit{minkowski}.
    \item \textbf{\textit{RF.}} Maximum depth of the tree: [2, 3, $\hdots$ , 9]. Number of trees: 100. Criterion: gini impurity.
    \item \textbf{\textit{AB.}} Number of estimators: [50, 100, 150]. Learning rate: [0.01, 0.1, 0.5, 1]. 
    \item \textbf{\textit{GB.}} Number of estimators: [50, 100, 150]. Learning rate: [0.01, 0.1, 0.5, 1]. Maximum depth of the individual estimators: [2, 4, 6, 8, 10]. 
\end{itemize}

Regarding the error metrics used to evaluate the performance of each of the four machine learning methods presented above, since we are dealing with a classification problem, the accuracy and the area under the ROC curve will be analyzed (AUC). Note that although the anonymization process is carried out on the complete database (in order to simulate the real case study), a stratified random train-test split (75$\%$-25$\%$) will be performed when training the models. 

\subsection{Results and analysis}

First, we will analyze the scenario in which the only anonymization technique applied is \textit{k-anonymity} for different values of $k$, and with a record suppression limit of 100$\%$ (i.e., there is no limit on the number of records that can be deleted to reach the anonymity condition). Specifically, the following values of $k$ will be taken: $k \in \{2, 5, 10, 25, 50, 75, 100\}$, although the analysis has been performed on more values of $k$, just the most significant ones are shown below. 

For this purpose, once the anonymization process with the hierarchies exposed in Subsection~\ref{sec:data} has been performed using the \textit{ARX Software}, we test the different machine learning models. Table~\ref{tab:acc_auc_varying_k} shows the results obtained both for the accuracy and for the AUC compared also to those obtained by applying the models on the raw data.

\begin{table}[ht]
    \centering
    \caption{Accuracy and AUC obtained for each machine learning model when varying the value of $k$ for $k$-anonymity.}
    \label{tab:acc_auc_varying_k}
    \resizebox{\linewidth}{!}{
    \begin{tabular}{c c@{\hspace{1.25\tabcolsep}}c c@{\hspace{1.5\tabcolsep}}c c@{\hspace{1.25\tabcolsep}}c  c@{\hspace{1.25\tabcolsep}}c}
    \toprule
        & \multicolumn{2}{c}{\textbf{\textit{kNN}}} & \multicolumn{2}{c}{\textbf{\textit{RF}}} & \multicolumn{2}{c}{\textbf{\textit{AB}}} & \multicolumn{2}{c}{\textbf{\textit{GB}}}\\
        \cmidrule{2-3}  \cmidrule{4-5}  \cmidrule{6-7}  \cmidrule{8-9}
        \textbf{\textit{k}} & \textit{Acc.} & \textit{AUC} & \textit{Acc.} & \textit{AUC} & \textit{Acc.} & \textit{AUC} & \textit{Acc.} & \textit{AUC}\\
        \midrule
         Raw & 0.8056 & 0.6779 & \textbf{0.8334} & \textbf{0.7351} & \textbf{0.8350} & 0.7377 & \textbf{0.8352} & \textbf{0.7437}\\
         2 & 0.8176 & 0.7143 & 0.8210 & 0.7215 & 0.8266 & 0.7372 & 0.8249 & 0.7346\\
         5 & \textbf{0.8199} & 0.7234 & 0.8208 & 0.7209 & 0.8197 & 0.7342 & 0.8221 & 0.7258\\
         10 & 0.8099 & 0.6932 & 0.8139 & 0.6913 &  0.8146 & 0.7171 & 0.8168 & 0.7157\\
         25 & 0.8170 & 0.7194 & 0.8163 & 0.6987 &  0.8164 & 0.7167 & 0.8177 & 0.7139\\
         50 & 0.8124 & \textbf{0.7268} & 0.8114 & 0.7285 & 0.8051 & \textbf{0.7399} & 0.8114 & 0.7285\\
         75 & 0.8129 & 0.7064 & 0.8129 & 0.7064 &  0.8091 & 0.7195 & 0.8129 & 0.7030\\
         100 & 0.8107 & 0.7254 & 0.8083 & 0.6981 & 0.8047 & 0.7117 & 0.8103 & 0.7274\\
         \bottomrule
    \end{tabular}}
\end{table}

Overall, starting first to analyze the results in terms of accuracy, we can observe that, as expected, a higher value of $k$ goes together with a reduction in accuracy in most of the cases. This is clearly seen in the three ensemble methods in which for both accuracy and AUC the best value is achieved with the raw data, and the worst with $k=100$ except for AUC with GB, which is obtained when $k=75$. However, with \textit{kNN} the minimum for the accuracy and the AUC is reached when using the raw data. This peculiarity can be attributed to the characteristics of the data, the dimensionality of the problem and the hierarchies applied. With respect to the AUC again, it can be observed that there is a lot of variability in the results, which start clearly decreasing for the first values of $k$ and the ensemble methods. It is especially striking that the optimal value for the AUC with \textit{AB} method is achieved when $k=50$ (although it is close to that obtained by using the raw data). 

Besides, we are also interested in analyzing for each of the previously exposed values of $k$, how close the resulting database ($\Omega_{k}$) is to being optimal for the stated level of anonymization. For this purpose the \textit{average equivalence class size metric}, which measures how well the equivalence classes are created to fit the best case \cite{ayala2014systematic}, is studied. Specifically, the optimal value for this metric would be one, indicating that all equivalence classes are of length $k$. Since in our example record suppression has been allowed with a limit of 100$\%$, instead of analyzing the number of equivalence classes as a function of the number of initial records (as exposed in the classic definition of this technique \cite{ayala2014systematic}), the number of records resulting from anonymization in each case will be considered. This is shown in Equation~\ref{eq:cavg}, in which \textit{ECs} is the set of equivalence classes. 

\begin{equation}
    C^{*}_{AVG_{k}}(\Omega_{k})=\frac{|\Omega_{k}|}{k|ECs|}
    \label{eq:cavg}
\end{equation}

As intuitively expected, the lowest value for this metric is reached for the highest value of $k$ analyzed ($k=100$), obtaining in particular the following values: $C^{*}_{AVG_{k}}=7.88, 10.94, 10.75, 7.68, 6.90, 5.13, 4.17$ $\forall k \in \{2,5,10,25,50,75,100\}$ respectively. 

\begin{figure}[ht]
    \centering
    \includegraphics[width = \linewidth]{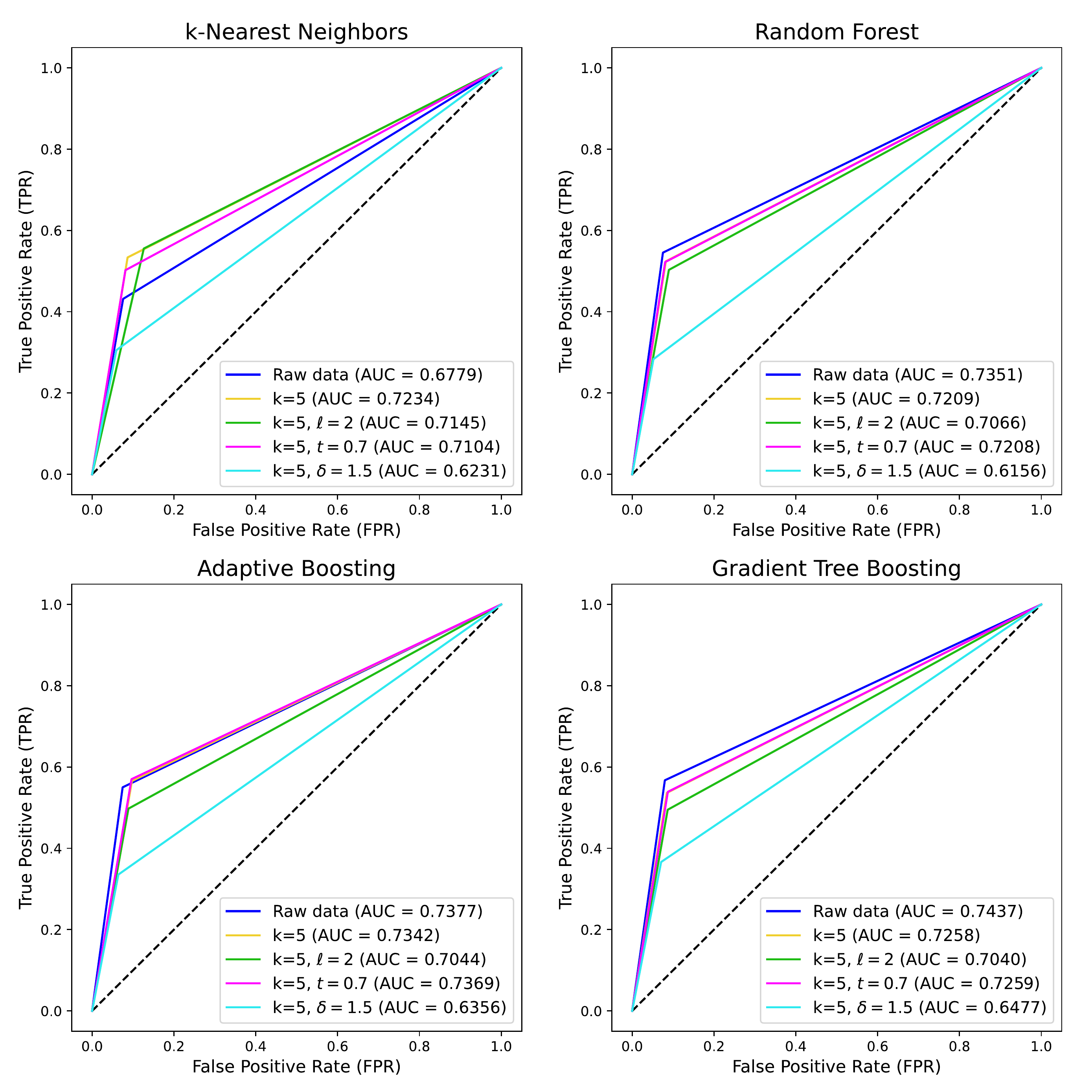}
    \caption{ROC curves and AUC obtained with each machine learning model and anonymity level for the \textit{adult dataset}.}
    \label{fig:roc_all}
\end{figure}

Next, once a value of $k=5$ for \textit{k-anonymity} has been fixed, the other three techniques described in Section~\ref{sec:anonymity} will be applied. In particular, a value $\ell=2$, $t=0.7$, and $\delta=1.5$ have been fixed in each case. Note that $\ell=2$ is the only possible value other than one, and for the other two cases, the values have been chosen after testing different values in order to achieve a balance with the number of deleted records and the utility. In addition, the values of $\delta$ has been chosen in order to be the most restrictive scenario of the five under study (this will be discussed further below in terms of the other parameters fulfilled). The results as a function of the accuracy are shown in Table~\ref{tab:accuracy}, and the ROC curves and the AUC obtained with each ML model analyzed and with each level of anonymity are displayed in Figure~\ref{fig:roc_all}. 
\begin{table}[ht]
    \centering
    \caption{Accuracy obtained with each machine learning model according to the anonymization technique applied.}
    \label{tab:accuracy}
    \begin{tabular}{cccccc}
         \toprule
          &  & & \textit{$k=5$,} & \textit{$k=5$,} & \textit{$k=5$,}\\
         \textbf{\textit{ML model}} & \textit{Raw} & \textit{$k=5$} & \textit{$\ell=2$} & \textit{$t=0.7$} & \textit{$\delta=1.5$}\\
         \midrule
         \textit{\textbf{kNN}} & 0.8056 & 0.8199 & 0.7894 & 0.8164 & 0.8061\\
         \textit{\textbf{RF}} & 0.8334 & 0.8208 & \textbf{0.8025} & \textbf{0.8218} & 0.8070\\
         \textit{\textbf{AB}} & 0.8350 & 0.8197 & 0.8017 & \textbf{0.8218} & 0.8085\\
         \textit{\textbf{GB}} & \textbf{0.8352} & \textbf{0.8221} & \textbf{0.8025} & 0.8210 & \textbf{0.8098}\\
         \bottomrule
    \end{tabular}
\end{table}

Here again the case of \textit{kNN} stands out for its counter-intuitiveness: both accuracy and AUC obtained by using the raw data are the second worst values of the five scenarios considered. However, with the ensemble methods the optimum is reached when using the models with the raw data (both for the accuracy and the AUC), i.e. without applying any anonymization technique. In the same line, both with \textit{kNN} and the ensemble models the worst results in terms of accuracy are obtained when the classification is performed with the data resulting after applying \textit{$\ell$-diversity} with $\ell=2$. This is really intuitive from the meaning of this property: for the same set of quasi-identifiers, the label will take in all cases two different values. 

Similarly, the worst performance in terms of AUC clearly occurs with all four models in the case where \textit{$\delta$-disclosure} is applied with $\delta = 1.5$. Again this is really intuitive if we analyze the parameters that are satisfied for each technique in that scenario, because although \textit{$\ell$-diversity} is only verified for $\ell=1$, the value of $t$ for \textit{t-closeness} is the most stringent of all the cases analyzed, being $t=0.47$. This has been verified by means of the Python library \textit{pyCANON} \cite{sainz2022python} (version 1.0.0.), and we can also observe that although in the case in which the value of $\delta$ has been set to 1.5, this property is verified for $\delta = 1.16$ (more restrictive). In the case where we set $\ell=2$, we obtain values $t=0.64$ and $\delta=4.88$, while in the case where we set $t=0.7$, $\ell=1$ and $\delta=4.10$

As expected, the three ensemble methods perform quite similarly in all scenarios, which is plainly illustrated in Figure~\ref{fig:roc_all}. In particular, the best value for the AUC is obtained with \textit{GB} and the raw data, and in the worst scenario ($\delta=1.5$) the best value is also obtained with this method for both accuracy (0.8098) and AUC (0.6477). Note that when $\delta=1.5$, although the prediction does not seem to be bad in terms of accuracy it is actually poor in view of the AUC with the four models. This reflects the need to test different error metrics in order to select an optimal anonymization technique and ML model.

In order to carry out a classification task, it should be noted that the optimum scenario would be for all the records that constitute an equivalence class to have the same label (i.e. the same sensitive attribute). However, this is in contrast to the three anonymization techniques analyzed that focus on sensitive attributes, since it would make homogeneity attacks feasible, among others. Let us therefore look at the \textit{classification metric (CM)} obtained in each of the four cases analyzed, defined as shown in Equation~\ref{eq:cm} (see \cite{10.1145/775047.775089}):

\begin{equation}
    CM = \frac{1}{N}\sum_{i=1}^{N}penalty(r_{i}),
    \label{eq:cm}
\end{equation}

where $penalty(r_{i})=1$ if the row $r_{i}$ has been deleted or if its associated label (i.e. SA) takes a value other than the majority value in the equivalence class to which it belongs, and 0 otherwise, $\forall i \in \{1,\hdots,N\}$ with $N$ the number of records in the original database.

The results obtained for each of the four cases and the CM metric are as follows: 0.2569 ($k=5$), 0.3089 ($k=5$, $\ell=2$), 0.2575 ($k=5$, $t=0.7$), 0.4589 ($k=5$, $\delta=1.5$). These values contrast with those obtained in Table~\ref{tab:accuracy} for the accuracy, where the results for $\delta=1.5$ are better than those for $t=0.7$ in 3 of the 4 cases, while they agree with those obtained for the AUC (see Figure~\ref{fig:roc_all}). Note that for the AUC the worst results are obtained when $\delta=1.5$, as would be expected based on the values calculated for \textit{CM}.

\section{Conclusions and future directions}\label{sec:conclusions}

In this paper the performance of four classical machine learning models in a classification task has been analyzed after subjecting the \textit{adult dataset} to different levels of anonymity. In particular, the scaling of the accuracy and the AUC has been analyzed when increasing the value of $k$ for \textit{k-anonymity}, as well as the optimally of the anonymization by using the $C^{*}_{AVG_{k}}$ metric exposed in Equation~\ref{eq:cavg}. As a result, it is noteworthy that in the case of \textit{kNN} model the best results in terms of both metrics are not obtained either with the raw data or with the lowest value of $k$. As for the ensemble methods, in all cases the best performance concerning accuracy are obtained when training with the raw data, as expected intuitively.

Furthermore, these same models have been analyzed when applying \textit{$\ell$-diversity}, \textit{t-closeness} and \textit{$\delta$-disclosure privacy} in addition to \textit{5-anonymity}, with $\ell=2$, $t=0.7$ and $\delta=1.5$ respectively. The AUC and the accuracy obtained in each case are studied, as well as the \textit{classification metric (CM)} defined in Equation~\ref{eq:cm}. In general terms in this case the best results are obtained when using the ensemble models, and it is observed that in agreement with \textit{CM} the worst results in terms of AUC are reached with $\delta=1.5$ in addition to $k=5$. 

As future lines to further extend this work, we are interested in the extrapolation to other datasets, anonymization techniques and their associated parameters. In addition, the inclusion of \textit{differential privacy (DP)} during the inference process using machine learning models is also an attractive field of study.

\section{Code availability}\label{sec:code}
For completeness and reproducibility of the work, the anonymized data together with the code carried out is available in the following GitHub repository: \url{https://github.com/IFCA/anonymity-ml}.

\section*{Acknowledgment}

The authors would like to thank the funding through the European Union - NextGenerationEU (Regulation EU 2020/2094), through CSIC's Global Health Platform (PTI+ Salud Global) and the support from the project AI4EOSC ``Artificial Intelligence for the European Open Science Cloud'' that has received funding from the European Union's Horizon Europe research and innovation programme under grant agreement number 101058593.


\end{document}